\documentclass[letterpaper, 10 pt, conference]{ieeeconf}  % Comment this line out if you need a4paper

\IEEEoverridecommandlockouts                              % This command is only needed if 
\usepackage{amsmath,amsfonts}
\usepackage{algorithmic}
\usepackage{algorithm}
\usepackage{array}

\usepackage[caption=false,font=normalsize,labelfont=sf,textfont=sf]{subfig}

\usepackage{textcomp}
\usepackage{stfloats}
\usepackage{url}
\usepackage{verbatim}
\usepackage{graphicx}
\usepackage{multirow} % 用于跨行
\usepackage{pifont} % 导入pifont宏包
\usepackage{colortbl} % 用于单元格颜色
\usepackage{xcolor}
\usepackage{booktabs}
\usepackage{hyperref}
\hypersetup{
	colorlinks=true,
	linkcolor=cyan,
	filecolor=blue,      
	urlcolor=black,
	citecolor=green,
}

\title{\LARGE \bf
Motion-Coupled Mapping Algorithm for Hybrid Rice Canopy
}

\author{Huaiqu Feng$^{1}$, Guoyang Zhao$^{2}$, Cheng Liu$^{2}$, Yongwei Wang$^{1,\dagger}$ and Jun Wang$^{1}$
\thanks{$^\dagger$ means corresponding author.}
\thanks{$^{1}$Huaiqu Feng, Yongwei Wang and Jun Wang are with the College of Biosystems Engineering and Food Science, Zhejiang University, China
(email: alaosion@zju.edu.cn.)}
\thanks{$^{2}$Guoyang Zhao and Cheng Liu are with the Robotics and Autonomous Systems Thrust, The Hong Kong University of Science and Technology (Guangzhou), China
(email:{\{gzhao492, cliu425\}@connect.hkust-gz.edu.cn).}}
}

\begin{document}

\maketitle
\thispagestyle{empty}
\pagestyle{empty}

%%%%%%%%%%%%%%%%%%%%%%%%%%%%%%%%%%%%%%%%%%%%%%%%%%%%%%%%%%%%%%%%%%%%%%%%%%%%%%%%
% \begin{abstract}
% This paper presents a motion-coupled mapping algorithm for Agricultural Unmanned Ground Vehicles (Agri-UGVs) operating in complex hybrid rice fields. Accurate canopy contour mapping is critical for efficient field navigation and task execution. Our algorithm integrates kinematic and inertial measurements with real-time RGB-D data to estimate the height and contour of the hybrid rice canopy. By refining probabilistic distributions of canopy morphology through grid mapping, the algorithm enhances both the perception and control of Agri-UGVs. The method has been tested on a high-clearance Agri-UGV platform, demonstrating reliable performance across diverse environments, including indoor tests and outdoor paddy fields. The results show significant improvements in mapping accuracy, contributing to more precise navigation and task efficiency in agricultural settings.
% \end{abstract}

\begin{abstract}
% This paper presents a motion-coupled mapping algorithm for contour mapping of hybrid rice canopies, specifically designed for Agricultural Unmanned Ground Vehicle (Agri-UGV) navigating complex and unknown rice fields. Precise canopy mapping is essential for Agri-UGV to plan efficient routes and avoid protected zones. The motion control of Agri-UGV, tasked with impurity removal and other operations, depends heavily on accurate estimation of rice canopy height and structure. To achieve this, the proposed algorithm integrates real-time RGB-D sensor data with kinematic and inertial measurements, enabling efficient mapping and proprioceptive localization. The algorithm produces grid-based elevation maps that reflect the probabilistic distribution of canopy contours, accounting for motion-induced uncertainties. It has been successfully implemented on a high-clearance Agri-UGV platform and tested in various environments, including both controlled and dynamic rice field settings. This approach significantly enhances the mapping accuracy and operational reliability of Agri-UGV, contributing to more efficient autonomous agricultural operations.

This paper presents a motion-coupled mapping algorithm for contour mapping of hybrid rice canopies, specifically designed for Agricultural Unmanned Ground Vehicles (Agri-UGV) navigating complex and unknown rice fields. Precise canopy mapping is essential for Agri-UGVs to plan efficient routes and avoid protected zones. The motion control of Agri-UGVs, tasked with impurity removal and other operations, depends heavily on accurate estimation of rice canopy height and structure. To achieve this, the proposed algorithm integrates real-time RGB-D sensor data with kinematic and inertial measurements, enabling efficient mapping and proprioceptive localization. The algorithm produces grid-based elevation maps that reflect the probabilistic distribution of canopy contours, accounting for motion-induced uncertainties. It is implemented on a high-clearance Agri-UGV platform and tested in various environments, including both controlled and dynamic rice field settings. This approach significantly enhances the mapping accuracy and operational reliability of Agri-UGVs, contributing to more efficient autonomous agricultural operations.
\end{abstract}

\section{Introduction}

Hybrid rice is a high-yield, high-quality, and stress-resistant variety of rice. As one of the major areas for hybrid rice cultivation, China accounts for approximately 160,000 hectares of hybrid rice planting area \cite{wei2022transcriptional}. To address the challenges of visual perception of rice canopies, it is essential to systematically study the progressive changes in the appearance of rice plants at different growth stages.

For the extraction of canopy height models, Unmanned Aerial Vehicles (UAV) have limited penetration, and point clouds mainly capture surface features of the canopy, especially in dense vegetation cover areas \cite{liu2018efficient}. It is difficult for UAV remote sensing to distinguish between canopy tops and ground-level points, particularly in dense crop fields \cite{lang2023high}.

\begin{figure}[t!]
    \centering
    \includegraphics[width=.45\textwidth]{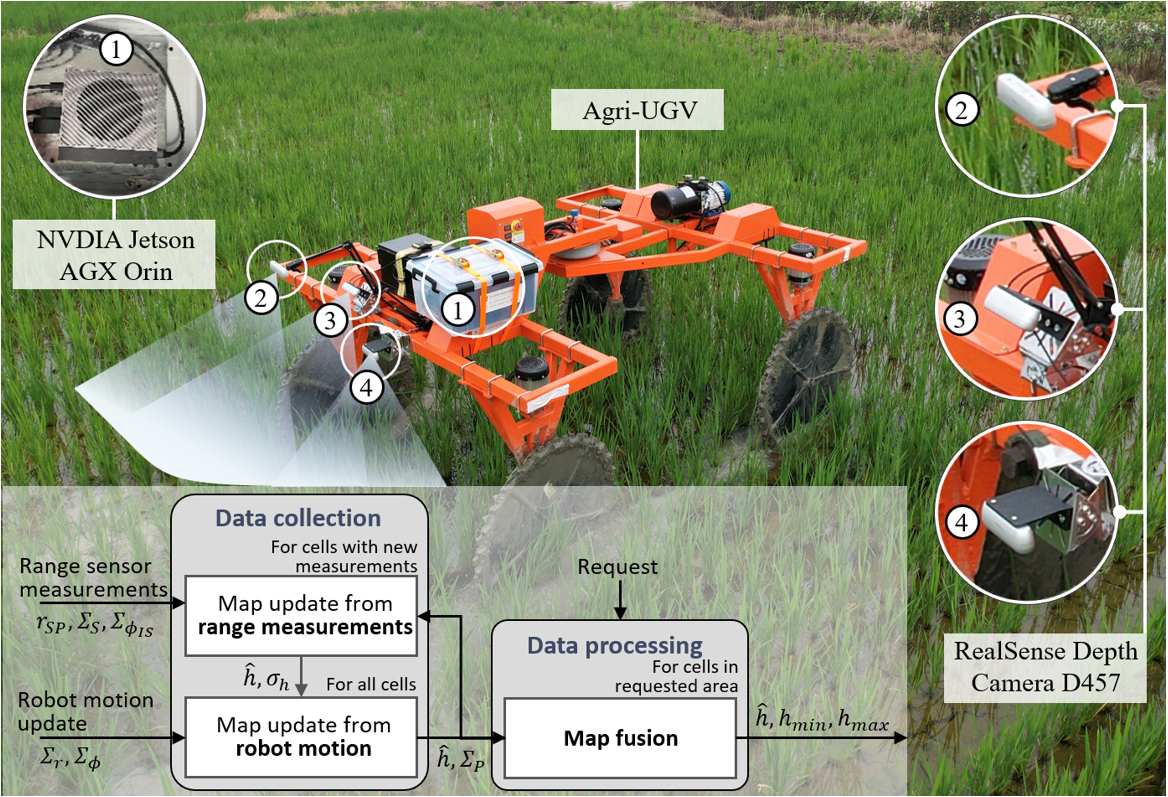}
    \vspace{-5pt}
    \caption{\textbf{Robot mapping in hybrid rice field.} We set up three different depth camera perspectives and use edge device for real-time computation.}
    \label{cover-fig}
    \vspace{-18pt}
\end{figure}

In terms of rice canopy perception, the complex and unstructured nature of rice fields presents significant challenges. These include background reflections from water layers, water surface glare, shadows, and interference from weeds. As rice plants grow denser after the closure of crop rows, factors such as small plant spacing, overlapping leaf canopies, varied leaf textures, and shading further complicate visual perception.

Accurate contour mapping of rice canopies is critical for Agricultural Unmanned Ground Vehicles (Agri-UGV) to efficiently plan operational areas and avoid protected zones \cite{wei2022row}. In the context of plant height detection, one of the primary advantages of mapping is its ability to precisely depict the physical structure and spatial arrangement of crops, including their height, width, and density. This is achieved through sensor data, such as depth sensors, LiDAR, or rangefinders, which capture the 3D structure of the scene. The 3D geometry of the environment is often represented using volumetric models like multi-body occupancy grids \cite{dryanovski2010multi}, truncated signed distance fields, or Euclidean signed distance fields \cite{oleynikova2017voxblox}. One of the most common methods in robotics for geometric mapping is the elevation map, where a segmented planar grid represents the terrain \cite{ewen2022these}, similar to surface reconstruction techniques.

To improve proprioceptive localization of Agri-UGVs, this paper introduces a motion-coupled mapping algorithm for hybrid rice canopy contour estimation, based on kinematic and inertial measurements. By leveraging real-time RGB-D data, the algorithm iteratively refines probabilistic estimates of canopy contours, accounting for drift and uncertainty in state estimation. A grid-mapping inference framework generates grid-based elevation maps with probabilistic contour estimates, including upper and lower confidence limits. The algorithm is evaluated using relative and absolute pose errors and operates in real-time within the ROS-Jetson system. It is deployed on a high-clearance Agri-UGV platform, successfully estimating the contour morphology of hybrid rice canopy height in various environments, including outdoor paddy fields.

\section{Methodology}

\subsection{Data Acquisition}
% The data collection site is located in Nanxun District, north of Huzhou City, within the Hangjia Plain single-season japonica rice area of Zhejiang Province, China. To enhance the visibility of the Undesired Variety, it has been artificially highlighted in blue. Fig. \ref{data} provides scenario diagrams with the front view displayed in panel (a) and the top view in panel (b). The width of the male rice plants is recorded as 30 mm, while the box width of both female rice plants and the chassis wheelbase is 1500 mm. During the working condition scenario, the sensors capture canopy data as the rice canopy is swept by the Agri-UGV.

\begin{figure}[t!]
    \centering
    \includegraphics[width=.45\textwidth]{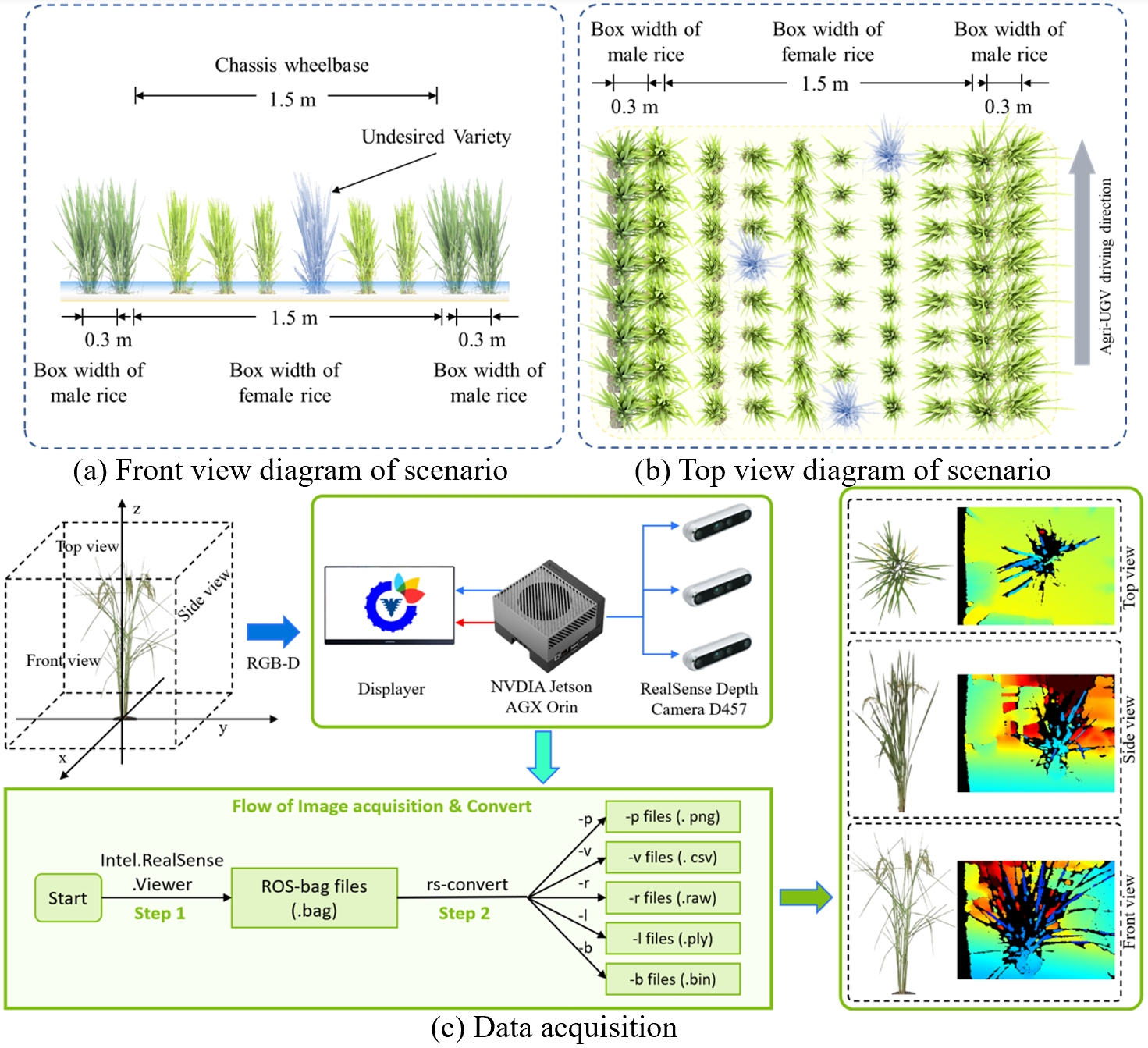}
    \vspace{-5pt}
    \caption{\textbf{Experimental scenario for data acquisition.}}
    \label{data}
    \vspace{-15pt}
\end{figure}

The data collection site is located in Nanxun District, north of Huzhou City, within the Hangjia Plain single-season japonica rice area of Zhejiang Province, China. To enhance the visibility of the undesired variety, it is artificially highlighted in blue. Fig. \ref{data} provides scenario diagrams with the front view displayed in panel (a) and the top view in panel (b). The width of the male rice plants is recorded as 30 mm, while both the female rice plants and the chassis wheelbase have a width of 1500 mm. During the operational scenario, the sensors capture canopy data as the Agri-UGV sweeps the rice canopy.

% The experimental setup used in this study includes an NVIDIA Jetson AGX Orin and three Intel RealSense D457 Depth Cameras. The Agri-UGV is composed of a vehicle body platform, an intelligent control system, and angle-adjustable stents. The vehicle body platform is powered by a built-in hub motor that enables constant speed and continuously variable speed adjustments. The Intel RealSense Depth Cameras are mounted on the angle-controlling stents, allowing for angle adjustment. The vertical shooting angle of the depth cameras can be adjusted from 0 to 90 degrees. (c) illustrates the information acquisition process and its working flow. Based on the field conditions observed during a prior survey, the camera spacing and the ground clearance of the Agri-UGV were optimized for better data collection.

The experimental setup includes an NVIDIA Jetson AGX Orin and three Intel RealSense D457 Depth Cameras. The Agri-UGV comprises a vehicle body platform, an intelligent control system, and angle-adjustable stents. The vehicle body platform is powered by a built-in hub motor that enables constant speed and continuously variable speed adjustments. The Intel RealSense Depth Cameras are mounted on the angle-controlling stents, allowing for angle adjustments from 0 to 90 degrees. Fig. \ref{data} (c) illustrates the information acquisition process and its workflow. Based on field conditions observed during a prior survey, the camera spacing and ground clearance of the Agri-UGV are optimized for better data collection.

\subsection{Overall Pipeline}
\begin{figure}[t!]
    \centering
    \includegraphics[width=.5\textwidth]{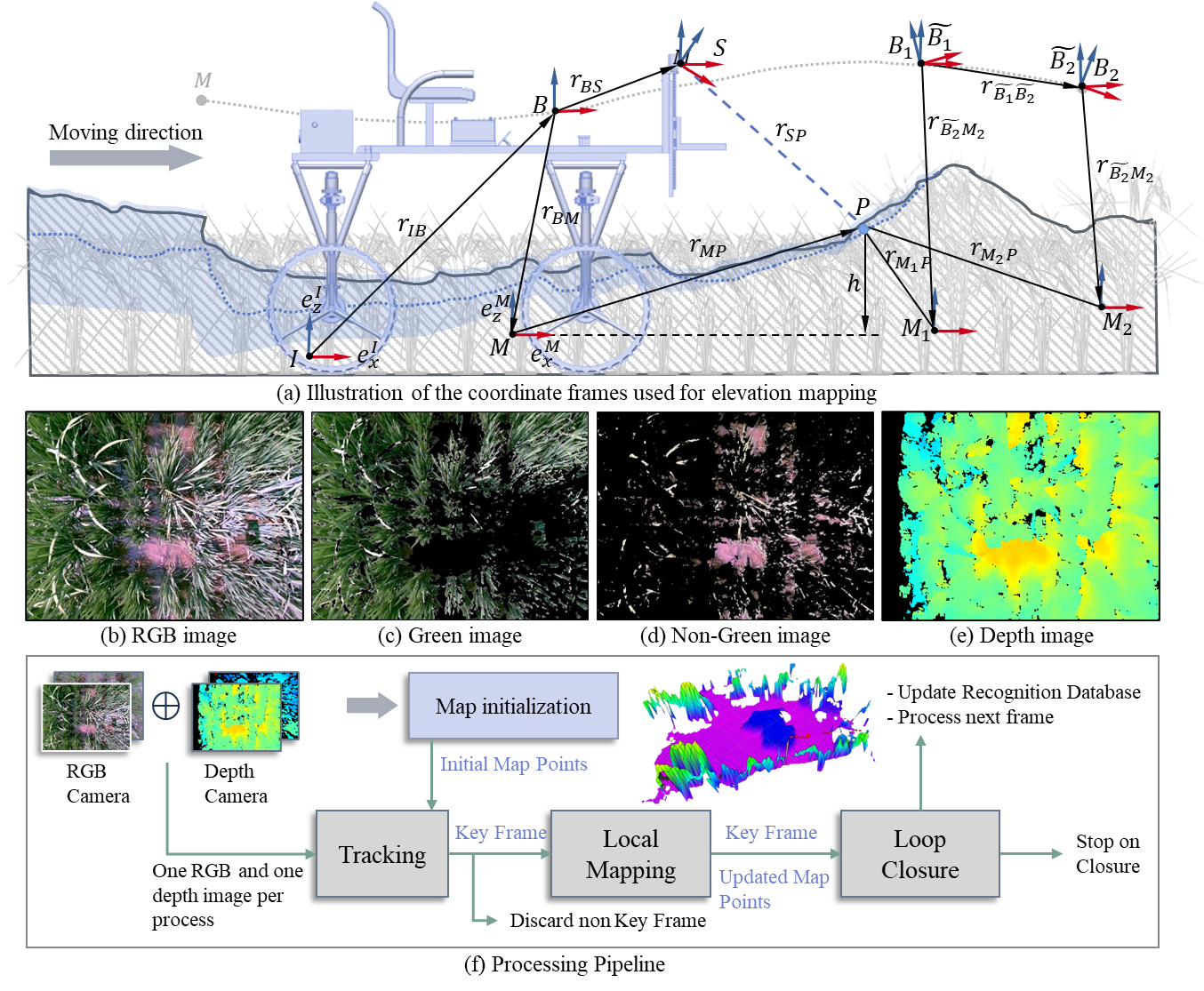}
    \vspace{-17pt}
    \caption{\textbf{Pipeline of hybrid rice canopy mapping.}}
    \label{pipeline}
    \vspace{-15pt}
\end{figure}

The Agri-UGV, equipped with depth cameras, pose estimators, and actuators, continuously scans and collects canopy information from the surrounding rice fields during movement. The hybrid rice canopy data, as shown in Fig. \ref{pipeline}(a), is captured by the depth cameras.
As depicted in Fig. \ref{pipeline}(b)-(e), the collected RGB-D data is pre-processed, where noise and discontinuities are removed using a Gaussian mixture model, resulting in a smoother and more accurate map.

The pipeline for RGB-D Visual Simultaneous Localization and Mapping (vSLAM), illustrated in Fig. \ref{pipeline}(e), is similar to monocular vSLAM \cite{mur2017visual,campos2021orb}. However, a key difference lies in the map initialization process: while monocular vSLAM generates 3D points from two color frames, RGB-D vSLAM utilizes both color and depth images to derive 3D map points.

The pipeline begins with Map Initialization, where 3D world points are constructed by extracting ORB (Oriented Fast and Rotated Brief) features from color images and computing their 3D locations using corresponding depth data. The first keyframe stores the initial color image and map points.
Next, during Tracking, the camera pose is estimated for each incoming RGB-D frame by matching features from the current color image to the most recent keyframe, ensuring continuous movement tracking.
In the Local Mapping stage, new 3D map points are computed from the depth image whenever a new keyframe is recognized. Once loop closure is detected, pose graph optimization is performed to refine the camera poses across all keyframes, leading to more accurate mapping and localization results.

\subsection{Coordinates Fusion}
\begin{figure}[t!]
    \centering
    \includegraphics[width=.5\textwidth]{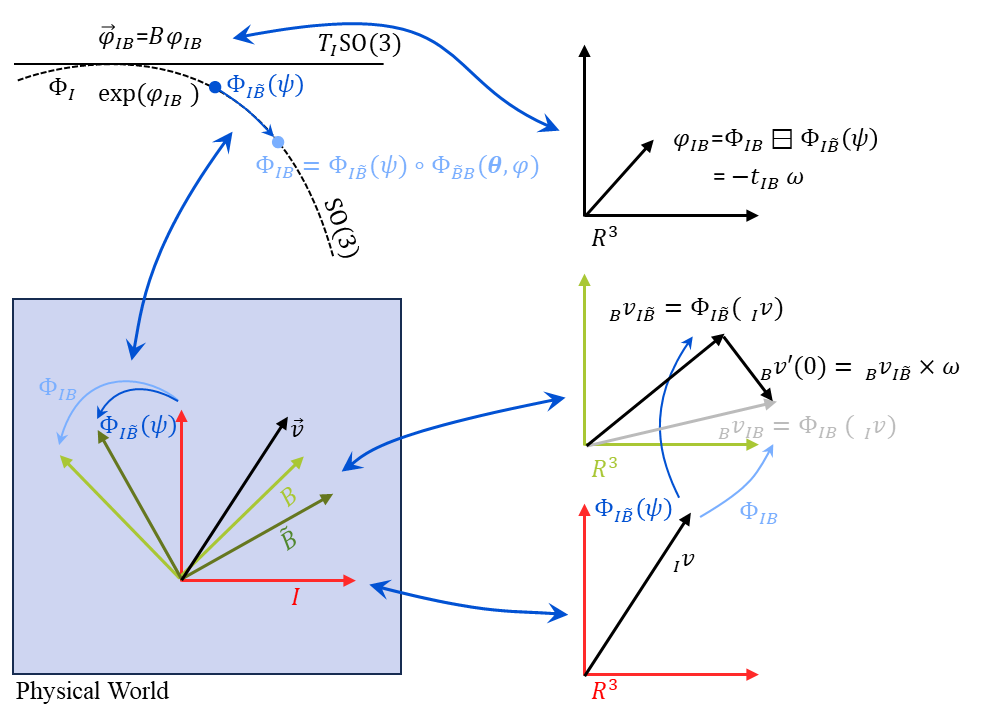}
    \vspace{-20pt}
    \caption{\textbf{Multiple coordinate fusion during robot motion.}}
    \label{coord}
    \vspace{-10pt}
\end{figure}

As shown in Fig. \ref{coord}, four coordinate frames are defined: the inertial frame \(I\), the Agri-UGV base frame \(B\), the RGB-D sensor frame \(S\), and the canopy map frame \(M\). The inertial frame \(I\) is fixed to the field environment, while the base frame \(B\) and sensor frame \(S\) are body-fixed frames attached to the Agri-UGV's center and the depth camera, respectively. The transformation between \(B\) and \(S\) is denoted by \((r_{BS}, \Phi_{BS})\), representing the uncertain spatial relationship between the Agri-UGV and the sensor \cite{cheeseman1987stochastic}. The relative orientation between the inertial frame \(I\) and the base frame \(B\) is given by \(\Phi_{IB} \in SO(3)\), as described by \cite{bloesch2016primer}.

The Agri-UGV's movement relative to the inertial frame is characterized by translation \(r_{IB}\) and rotation \(\Phi_{IB}\), which are estimated by the pose estimation system. This transformation is represented by a six-dimensional pose covariance matrix \(\Sigma_{IB} = \text{cov}(r_{IB}, \Phi_{IB})\). The rotation \(\Phi_{IB}\) can be further decomposed into yaw, pitch, and roll angles as follows:
% \[
% \Phi_{IB} = \Phi_{\tilde{IB}}(\psi) \circ \Phi_{\tilde{B}B}(\theta, \phi)
% \]
\begin{equation}
\Phi_{IB} = \Phi_{\tilde{IB}}(\psi) \circ \Phi_{\tilde{B}B}(\theta, \phi)
\end{equation}

where \(\Phi_{\tilde{IB}}(\psi)\) describes the rotation around the vertical axis with yaw angle \(\psi\), and \(\Phi_{\tilde{B}B}(\theta, \phi)\) accounts for the tilt between frames \(B\) and \(\tilde{B}\) with pitch \(\theta\) and roll \(\phi\).

The canopy map frame \(M\) is aligned with the inertial frame along the vertical axis, ensuring that \(e_z^I = e_z^M\), and its yaw angle \(\psi\) matches the yaw of the Agri-UGV base frame. The 3D position of a grid cell \(i\) in the map is given by \(P_i = (x_i, y_i, \hat{h}_i)\), where \(x_i\) and \(y_i\) represent the grid position, and \(\hat{h}_i\) is the estimated height of the terrain.

The RGB-D sensor frame \(S\) is statically linked to the Agri-UGV base frame \(B\), while the canopy map frame \(M\) is defined relative to \(B\). Points captured in the RGB-D sensor frame are transformed into the canopy map frame. The semantic mapping node converts elevation maps into semantic maps, categorizing elevation values as corresponding hybrid rice canopy contours.

The algorithm estimates the canopy's morphological attributes by processing RGB-D data to extract features such as normal vectors, gradients, and curvature. A Gaussian convolution method smooths the elevation data, allowing for recursive mapping of each pixel's elevation, where each pixel records the plant height.

\subsection{Map Update from Range Measurements}
When updating the canopy elevation map from range measurements, the height data collected by the RGB-D sensor is converted into the Agri-robot’s coordinate system. These measurements are treated as spatial points and mapped onto the elevation grid. Each measurement at a grid cell \((x, y)\) is represented as a new hybrid rice height measurement, denoted by \( \tilde{p} \sim N(p, \sigma_p^2) \), where \(p\) is the mean height and \(\sigma_p^2\) the variance.

The height measurement \(p\), obtained in the RGB-D sensor frame \(S\), is converted to the canopy map frame \(M\) using the following relation:
\begin{equation}
p = P\left(\Phi_{SM}^{-1} (\mathbf{r}_{SP}) - \mathbf{r}_{SM}\right)
\end{equation}

% \[
% p = P(\Phi_{SM}^{-1} (\mathbf{r}_{SP}) - \mathbf{r}_{SM})
% \]

Based on the Agri-robot's pose, the height data is matched with the global elevation map to update the terrain model. The difference between the new measurement and the corresponding height in the global map is used to refine the hybrid rice canopy contour. The projection matrix \(P = \begin{bmatrix} 0 & 0 & 1 \end{bmatrix}\) reduces the 3D measurement to a single height measurement in the map frame.

To calculate the variance \(\sigma_p^2\), the Jacobians of the sensor measurement \(J_S\) and rotation \(J_\Phi\) are computed as:
\begin{equation}
\begin{aligned}
J_S &= \frac{\partial p}{\partial \mathbf{r}_{SP}} = P C(\Phi_{SM})^\top \\
J_\Phi &= \frac{\partial p}{\partial \Phi_{SM}} = P C(\Phi_{SM})^\top \mathbf{r}_{SP}^\times
\end{aligned}
\end{equation}
% \[
% J_S = \frac{\partial p}{\partial \mathbf{r}_{SP}} = P C(\Phi_{SM})^\top
% \]
% \[
% J_\Phi = \frac{\partial p}{\partial \Phi_{SM}} = P C(\Phi_{SM})^\top \mathbf{r}_{SP}^\times
% \]

The error propagation for the variance is given by:
\begin{equation}
\sigma_p^2 = J_S \Sigma_S J_S^\top + J_\Phi \Sigma_{\Phi_{IS}} J_\Phi^\top
\end{equation}
% \[
% \sigma_p^2 = J_S \Sigma_S J_S^\top + J_\Phi \Sigma_{\Phi_{IS}} J_\Phi^\top
% \]

Here, \(\Sigma_S\) is the covariance of the range sensor, and \(\Sigma_{\Phi_{IS}}\) is the covariance of the sensor's rotation.

The new height measurement \((\tilde{p}, \sigma_p^2)\) is merged with the current map estimate \((\tilde{h}, \sigma_h^2)\) using a 1D Kalman filter:
\begin{equation}
\hat{h}^+ = \frac{\sigma_p^2 \hat{h}^- + \sigma_h^{2-} \tilde{p}}{\sigma_p^2 + \sigma_h^{2-}}, \quad
\sigma_h^{2+} = \frac{\sigma_h^{2-} \sigma_p^2}{\sigma_h^{2-} + \sigma_p^2}
\end{equation}
% \[
% \hat{h}^+ = \frac{\sigma_p^2 \hat{h}^- + \sigma_h^{2-} \tilde{p}}{\sigma_p^2 + \sigma_h^{2-}}, \quad
% \sigma_h^{2+} = \frac{\sigma_h^{2-} \sigma_p^2}{\sigma_h^{2-} + \sigma_p^2}
% \]

For multiple measurements within the same grid cell, the fusion is based on Mahalanobis distance, retaining the highest elevation measurement and discarding outliers.

To improve localization, the algorithm uses RGB-D data to perform feature matching between local and global maps. By extracting feature descriptions from the local contour map and comparing them with the global elevation map, the system aligns the local map with the global one, updating both the Agri-robot's position and the global elevation map.

% \section{EXPERIMENTS}
% \subsection{Implementation}
% The algorithm was implemented on the high-performance Jetson AGX Orin platform, running Ubuntu 20.04 to ensure system stability and compatibility. ROS Noetic was selected as the primary development framework, providing a comprehensive, flexible, and feature-rich environment for the Robot Operating System (ROS).

% A complete ROS node framework was established, covering key functionalities such as sensor data processing, state estimation, loop closure detection, visualization, device calibration, and canopy mapping. Using the D457 depth camera, combined with grid mapping techniques and the VINS-Fusion-gpu system \cite{qin2018vins}, high-precision state estimation and 3D environment reconstruction were achieved.

\section{EXPERIMENTS}
\subsection{Implementation}
The algorithm is implemented on the high-performance Jetson AGX Orin platform, running Ubuntu 20.04 to ensure system stability and compatibility. ROS Noetic is selected as the primary development framework, providing a comprehensive, flexible, and feature-rich environment for the Robot Operating System (ROS).

A complete ROS node framework is established, covering key functionalities such as sensor data processing, state estimation, loop closure detection, visualization, device calibration, and canopy mapping. Using the D457 depth camera, combined with grid mapping techniques and the VINS-Fusion-gpu system \cite{qin2018vins}, high-precision state estimation and 3D environment reconstruction are achieved.

% \begin{table}[h]
% \centering
% \caption{VINS-Fusion Parameter Settings}
% \begin{tabular}{|l|c|}
% \hline
% \textbf{Parameter}           & \textbf{Value}       \\ \hline
% multiple\_thread             & 1                    \\ \hline
% max\_cnt                     & 150                  \\ \hline
% min\_dist                    & 30                   \\ \hline
% frequency                    & 10Hz                 \\ \hline
% F\_threshold                 & 1.0 pixel            \\ \hline
% show\_track                  & 1                    \\ \hline
% flow\_back                   & 1                    \\ \hline
% max\_solver\_time            & 0.04 ms              \\ \hline
% max\_num\_iterations         & 8                    \\ \hline
% keyframe\_parallax           & 10.0 pixels          \\ \hline
% acc\_n                       & 0.04                 \\ \hline
% gyr\_n                       & 0.004                \\ \hline
% acc\_w                       & 0.002                \\ \hline
% gyr\_w                       & 4.0e-5               \\ \hline
% g\_norm                      & 9.805 m/s²           \\ \hline
% estimate\_td                 & 1                    \\ \hline
% td                           & 0.00 s               \\ \hline
% \end{tabular}
% \label{table:VINS-Fusion-params}
% \end{table}

\begin{table}[h]
\centering
\caption{VINS-Fusion Parameter Settings}
\vspace{-3pt}
\begin{tabular}{l c | l c}
\toprule
\textbf{Parameter}           & \textbf{Value}       & \textbf{Parameter}           & \textbf{Value}       \\ \midrule
multiple\_thread             & 1                    & max\_cnt                     & 150                  \\ 
min\_dist                    & 30                   & frequency                    & 10Hz                 \\ 
F\_threshold                 & 1.0 pixel            & show\_track                  & 1                    \\ 
flow\_back                   & 1                    & max\_solver\_time            & 0.04 ms              \\ 
max\_num\_iterations         & 8                    & keyframe\_parallax           & 10.0 pixels          \\ 
acc\_n                       & 0.04                 & gyr\_n                       & 0.004                \\ 
acc\_w                       & 0.002                & gyr\_w                       & 4.0e-5               \\ 
g\_norm                      & 9.805 m/s²           & estimate\_td                 & 1                    \\ 
\bottomrule
\end{tabular}
\label{table:VINS-Fusion-params}
\vspace{-10pt}
\end{table}

\subsection{Mapping Results in Complex Environment}
\begin{figure}[t!]
    \centering
    \includegraphics[width=.46\textwidth]{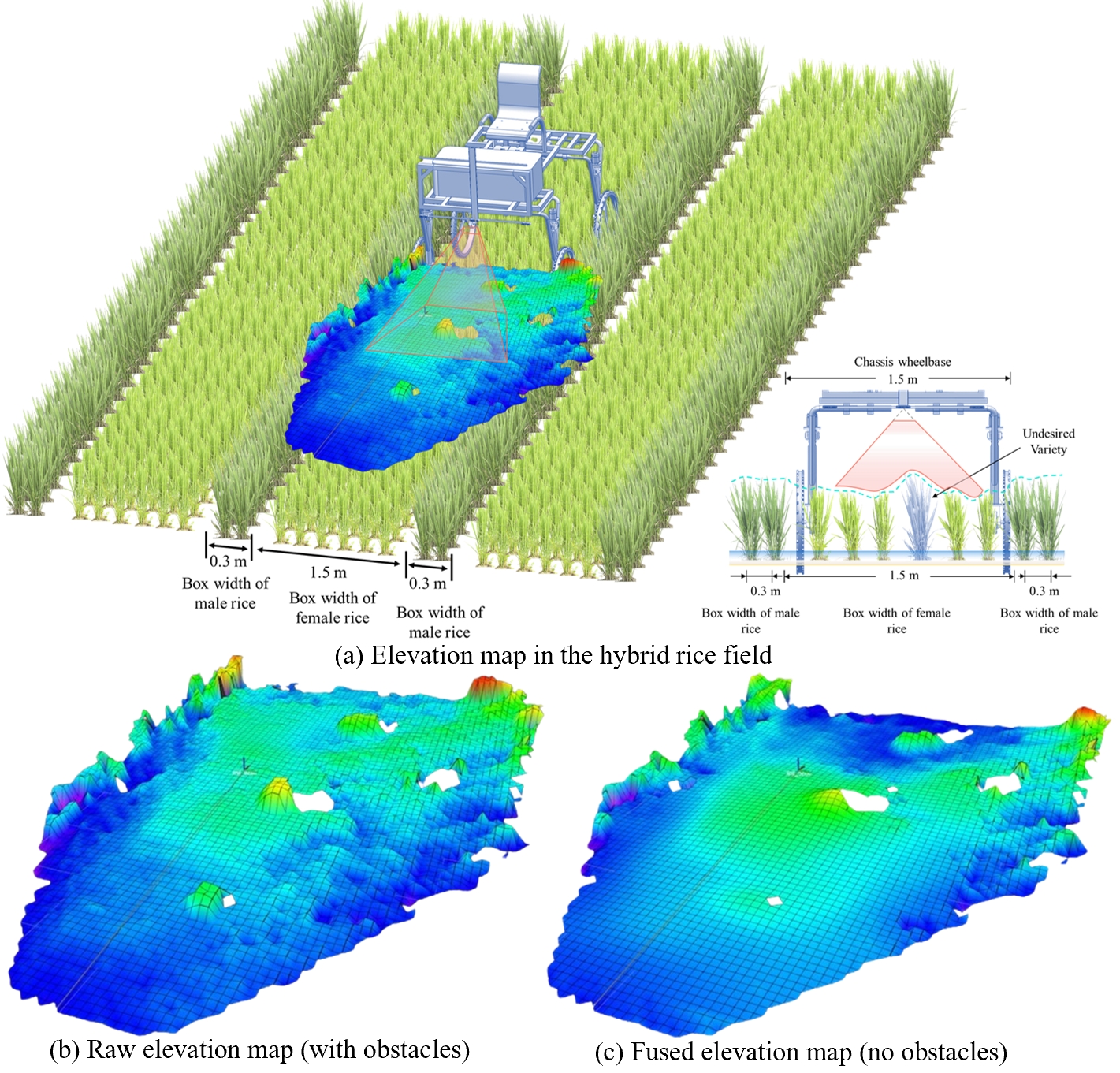}
    \vspace{-8pt}
    \caption{\textbf{Canopy mapping in hybrid rice field.}}
    \label{result1}
    \vspace{-10pt}
\end{figure}

Fig. \ref{result1} shows the real-time grid mapping results of the hybrid rice canopy contour. In (a), a schematic of the canopy mapping under uncertain conditions after the Agri-UGV's movement is illustrated. The raw elevation map in (b) reveals traces of previous obstacles in the hybrid rice field. However, the fused elevation map in (c) smooths these obstacles, creating a more seamless transition.

The Agri-UGV, equipped with three RGB-D sensors, generates a point cloud, which is converted into a canopy map of the hybrid rice. Both the point cloud and odometry messages are published at a frequency of 10 Hz. As the Agri-UGV moves, the canopy contour map is continuously updated and expanded based on the point cloud data. When the entire map is updated, including areas beyond the current view of the Agri-UGV, obstacles that were previously visible are often smoothed out, though they may temporarily reappear. The current view overlaps with the point cloud, while older map sections are constantly updated, even without new points being published in the point cloud.

Attempts to improve the visibility of canopy mapping do not prevent the smoothing of obstacles outside the Agri-UGV's current view. This smoothing effect is likely due to increasing uncertainty in the system. As uncertainty accumulates, particularly in areas behind the Agri-UGV, obstacles may disappear and reappear. The growing uncertainty is influenced by the RGB-D sensor noise model and the Agri-UGV's pose covariance, which increases with distance from the vehicle.

% \subsection{Canopy Contour Map and Path}
% In the complex growth environment of hybrid rice, a complete canopy contour map was successfully generated using the Agri-UGV platform, upon which consistent path trajectories were established, as shown in Fig. \ref{result2}. (a) illustrates the simulated trajectory of the Agri-UGV in a hybrid rice environment, while (b) presents images of the Agri-UGV navigating through the dense rice canopy at a speed of 1 m/s during outdoor field tests.

% The positioning performance of the RGB-D sensor module in the canopy mapping task was evaluated across three hybrid rice paddies. Real-time data collected from the RGB-D sensor was processed by the canopy mapping algorithm, producing a consistent and optimized map. It is important to note that the algorithm operates in localization mode and does not attempt to build a global map \cite{hess2016real}. The focus of the evaluation was to quantify performance degradation in real-time localization without trajectory bundle adjustment. The Agri-UGV's position estimates were compared against pseudo-ground-truth trajectories generated from trajectory bundles within pre-existing maps.

% Due to slippage and subsidence caused by muddy terrain and the weight of the Agri-UGV during movement, it is challenging to obtain accurate ground truth for the hybrid rice canopy elevation, making strict quantitative analysis difficult. Additionally, the slender tips of rice leaves posed challenges for precise detection in the canopy map, as they were often missed by the sensors.

\subsection{Canopy Contour Map and Path}
\begin{figure}[t!]
    \centering
    \includegraphics[width=.5\textwidth]{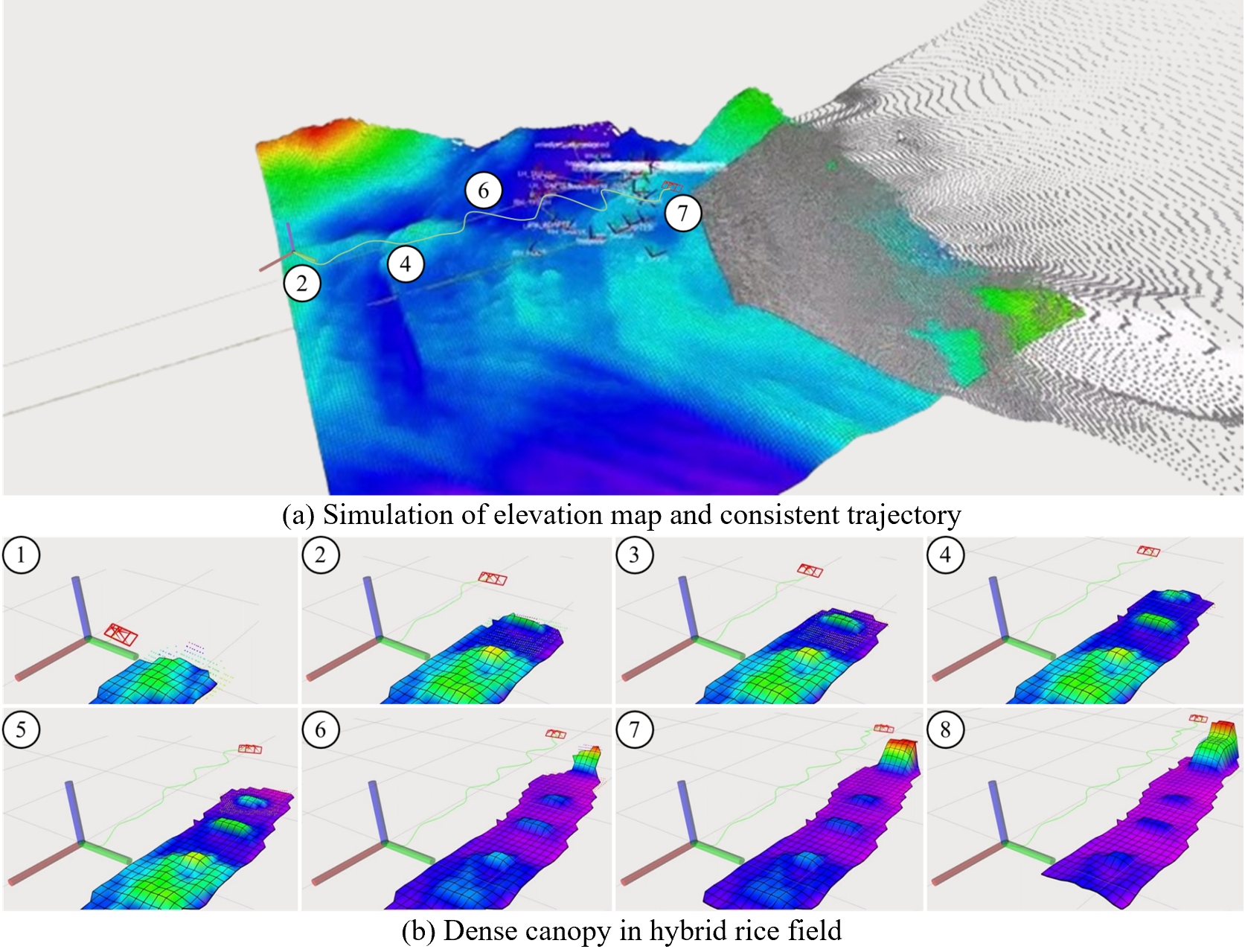}
    \vspace{-20pt}
    \caption{\textbf{Consistent map and trajectory in complex hybrid rice field.}}
    \label{result2}
    \vspace{-10pt}
\end{figure}

In the complex growth environment of hybrid rice, a complete canopy contour map is successfully generated using the Agri-UGV platform, upon which consistent path trajectories are established, as shown in Fig. \ref{result2}. (a) illustrates the simulated trajectory of the Agri-UGV in a hybrid rice environment, while (b) presents images of the Agri-UGV navigating through the dense rice canopy at a speed of 1 m/s during outdoor field tests.

The positioning performance of the RGB-D sensor module in the canopy mapping task is evaluated across three hybrid rice paddies. Real-time data collected from the RGB-D sensor is processed by the canopy mapping algorithm, producing a consistent and optimized map. It is important to note that the algorithm operates in localization mode and does not attempt to build a global map \cite{hess2016real}. The focus of the evaluation is to quantify performance degradation in real-time localization without trajectory bundle adjustment. The Agri-UGV's position estimates are compared against pseudo-ground-truth trajectories generated from trajectory bundles within pre-existing maps.

Due to slippage and subsidence caused by muddy terrain and the weight of the Agri-UGV during movement, it is challenging to obtain accurate ground truth for the hybrid rice canopy elevation, making strict quantitative analysis difficult. Additionally, the slender tips of rice leaves pose challenges for precise detection in the canopy map, as they are often missed by the sensors.

% \section{Conclusion}
% This research has demonstrated the importance of accurate rice canopy contour mapping for the efficient operation of Agri-UGV. The proposed algorithm, utilizing real-time RGB-D depth sensor data, effectively estimates the height and canopy contour of hybrid rice. Successfully implemented on the ROS-Jetson system and tested on a High-clearance Agri-UGV platform, the algorithm has shown precise localization and mapping capabilities in outdoor environments. This work significantly advances autonomous agricultural practices by providing a reliable solution for Agri-UGV to perform tasks such as impurity removal of heterologous varieties in complex rice farming settings. Future research will focus on further refining the algorithm for broader applications and improving its performance in diverse agricultural environments.

\section{Conclusion}
This research demonstrates the importance of accurate rice canopy contour mapping for the efficient operation of Agri-UGV. The proposed algorithm, utilizing real-time RGB-D depth sensor data, effectively estimates the height and canopy contour of hybrid rice. Successfully implemented on the ROS-Jetson system and tested on a high-clearance Agri-UGV platform, the algorithm shows precise localization and mapping capabilities in outdoor environments. This work significantly advances autonomous agricultural practices by providing a reliable solution for Agri-UGV to perform tasks such as impurity removal of heterologous varieties in complex rice farming settings. Future research focuses on further refining the algorithm for broader applications and improving its performance in diverse agricultural environments.


%%%%%%%%%%%%%%%%%%%%%%%%%%%%%%%%%%%%%%%%%%%%%%%%%%%%%%%%%%%%%%%%%%%%%%%%%%%%%%%%

%%%%%%%%%%%%%%%%%%%%%%%%%%%%%%%%%%%%%%%%%%%%%%%%%%%%%%%%%%%%%%%%%%%%%%%%%%%%%%%%



% \section*{ACKNOWLEDGMENT}

% The preferred spelling of the word ÒacknowledgmentÓ in America is without an ÒeÓ after the ÒgÓ. Avoid the stilted expression, ÒOne of us (R. B. G.) thanks . . .Ó  Instead, try ÒR. B. G. thanksÓ. Put sponsor acknowledgments in the unnumbered footnote on the first page.

%%%%%%%%%%%%%%%%%%%%%%%%%%%%%%%%%%%%%%%%%%%%%%%%%%%%%%%%%%%%%%%%%%%%%%%%%%%%%%%%

\clearpage
% \normalem
% \renewcommand*{\bibfont}{\footnotesize}
% \printbibliography

\bibliographystyle{ieeetr} % 使用IEEE的样式
\bibliography{ref}         % 引用你的BibTeX文件

\end{document}